\documentclass{article}

\usepackage[square,comma,sort&compress,numbers]{natbib}

\usepackage[preprint]{neurips_2021}

\usepackage[utf8]{inputenc} 
\usepackage[T1]{fontenc}    
\usepackage{hyperref}       
\usepackage{url}            
\usepackage{booktabs}       
\usepackage{amsfonts}       
\usepackage{nicefrac}       
\usepackage{microtype}      
\usepackage{xcolor}         
\usepackage{bm}
\usepackage{comment}
\usepackage{pgfplots}
\usepackage{amsmath,amssymb}

\title{Parameter Estimation for the SEIR Model\\Using Recurrent Nets}

\author{
Chun Fan$^{\bigstar\blacktriangle}$,
Yuxian Meng$^\clubsuit$,
Xiaofei Sun$^\clubsuit$
Fei Wu$^\blacklozenge$,
Tianwei Zhang$^\spadesuit$,
Jiwei Li$^{\blacklozenge\clubsuit}$\\
$^\clubsuit$ Shannon.AI, $^\bigstar$Peking University, $^\blacktriangle$Peng Cheng Laboratory \\
$^\blacklozenge$Zhejiang University, $^\spadesuit$Nanyang Technological University \\ 
  \{yuxian\_meng, xiaofei\_sun, jiwei\_li\}@shannonai.com\\
  fanchun@pku.edu.cn, wufei@zju.edu.cn, tianwei.zhang@ntu.edu.sg
}

\begin{document}

\maketitle

\begin{abstract}
The standard way to estimate the parameters $\Theta_\text{SEIR}$ (e.g., the transmission rate $\beta$) of an SEIR model is to use grid search, where simulations are performed on each set of parameters, and the parameter set leading to the least $L_2$ distance between predicted number of infections and observed infections is selected. This brute-force strategy is not only time consuming, as simulations are slow when the population is large, but also inaccurate, since it is impossible to enumerate all parameter combinations. To address these issues, in this paper, we propose to transform the  non-differentiable problem of finding optimal $\Theta_\text{SEIR}$ to a differentiable one, where we first train a recurrent net  to fit a small number of simulation data. Next, based on this recurrent net that is able to generalize SEIR simulations, we are able to transform the objective to a differentiable one with respect to $\Theta_\text{SEIR}$, and straightforwardly obtain its optimal value. The proposed strategy is both time efficient as it only relies on a small number of SEIR simulations, and accurate as we are able to find the optimal $\Theta_\text{SEIR}$ based on the differentiable objective. On two COVID-19 datasets, we observe that the proposed strategy leads to significantly better parameter estimations with a smaller number of simulations.  
\end{abstract}

\section{Introduction}
The SEIR model \cite{kermack1927contribution,bjornstad2002dynamics,harko2014exact,kendall2020deterministic} is a widely-used epidemiological model to predict the macroscopic behavior of disease spread through a population, e.g., the spread of COVID-19 among different populations at different locations over time \cite{estrada2020COVID,prem2020effect,godio2020seir,aleta2020modelling,chang2021mobility,grimm2021extensions}. A typical SEIR model  models the spread dynamics of disease 
using four states of populations: susceptible (S), exposed (E), infectious (I) and recovered (R), where susceptible individuals can be transformed into the exposed, and later to the infectious and finally to the recovered. 
 Each  transition is associated with corresponding parameter(s), forming the parameter set  $\Theta_\text{SEIR}$ for the model: 
{\it transmission rate} for $S\rightarrow E$,
 {\it infection rate}  for $E\rightarrow I$ and 
 {\it recovery rate} for $I\rightarrow R$.
 $\Theta_\text{SEIR}$ captures the spreading pattern of the disease, 
 the learning of which is thus crucial for  understanding the disease spread, and developing  
 social-distancing or 
interventions policies. 

Existing SEIR models  relies on grid search to estimate 
$\Theta_\text{SEIR}$ 
 via simulations. Model simulations are performed on different sets of parameters, and the set that has the smallest $L_2$ distance between the predicted number of infections and the observed number of infections is selected as the best parameter combination. 
This is because of the non-differentiable nature of the $L2$ objective with respect to $\Theta_\text{SEIR}$:
predicted number of infections are obtained from discrete simulations. 
This learning process is 
(1) {\it high time-intensive}: the simulation process can be slow when the population is large, which gives a time complexity of $\mathcal{O}(MNT)$ where $M$ is the number of simulations performed, $N$ is total number of individuals and $T$ is maximum time steps; besides, the search space will grow exponentially with respect to the number of parameters we need to estimate;
and (2) {\it inaccurate}: because it is impractical to enumerate all possible combinations of parameters in the continuous space, the resulting parameter combination would be sub-optimal given a limited amount of trials.

To address these issues, in this paper, we propose to transform the original non-differentiable simulation problem into a differentiable one using neural recurrent nets for parameter estimation in SEIR models.
The basic idea is that neural recurrent nets are differentiable with respect to the input learnable parameters, so that they are able to automatically learn these parameters via gradient descent and backpropagation \cite{rumelhart1986learning}.
Another advantage neural recurrent nets offer is that they intrinsically model the temporal changes of observations, and have the potentials to make accurate predictions about the number of individuals at each time step. 
To train the recurrent net, we first harvest the training data by running a few SEIR simulations, leading to a collection of simulated data under different parameter combinations. Next, a recurrent net is trained to fit these simulations. Then with model parameters of the the recurrent net $\Theta_\text{LSTM}$ fixed, the observation data are used to train the net with respect to the SEIR parameters $\Theta_\text{SEIR}$ via gradient descent, leading to the optimal parameter values $\Theta^*_\text{SEIR}$.

The proposed strategy is both time efficient in that it only relies on a small number of SEIR simulations, and accurate because we are able to find the optimal parameters $\Theta^*_\text{SEIR}$ based on the differentiable objective. 
Through experiments on two COVID-19 datasets, we observe that the proposed strategy leads to more accurate parameter estimates with significantly better  time efficiency.

\section{Background and Problem Statement}
\begin{figure}[t]
  \centering
  \includegraphics[scale=0.8]{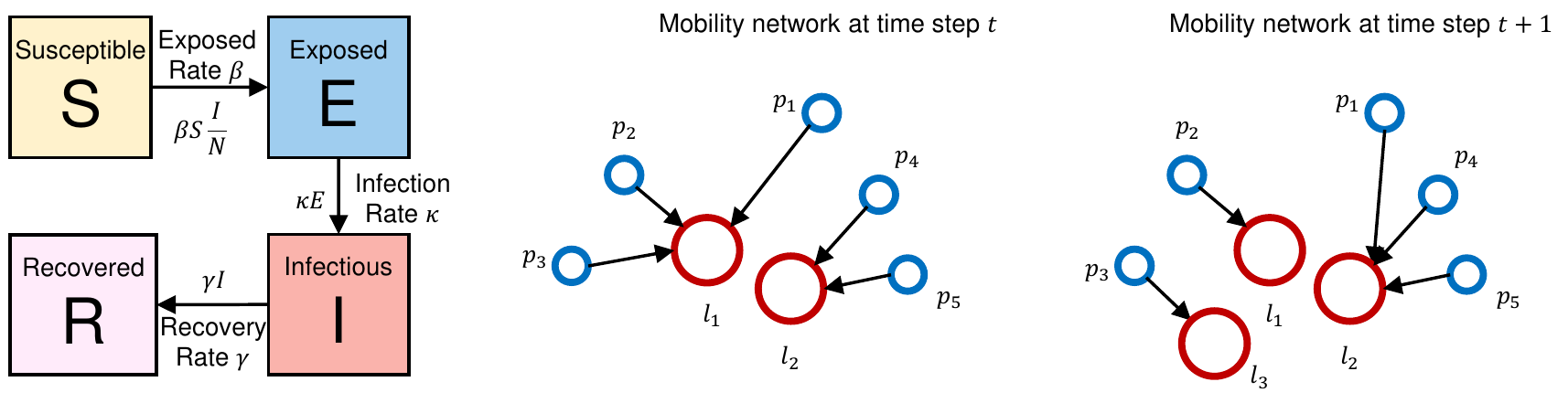}
  \caption{An overview of a standard SEIR model (left) the mobility network (right: from time $t$ to time $t+1$, person $p_3$ moved from location $l_1$ to $l_3$, and $p_1$ moved from $l_1$ to $l_2$).}
  \label{fig:seir}
\end{figure}

This work focuses on estimating the parameters $\Theta_\text{SEIR}$ in the SEIR mode. A standard SEIR model is formulated in terms of 4 populations of individuals: the susceptible population ($S$), the exposed population ($E$), the infectious population ($I$) and the recovered population ($R$), which are respectively comprised of all individuals susceptible to the infection,all individuals that have contacted infected patients but are currently during their incubation period, infected individuals that can transmit the disease to susceptible population, and recovered individuals that cannot become infected again and cannot transmit the disease to others.
Given a mobility network $\mathcal{G}_t=(\{\mathcal{P},\mathcal{L}\},\mathcal{E})_t$ at time step $t (t=1,2,\cdots,T)$ where $\mathcal{P}$ is the set of person nodes, $\mathcal{L}$ is the set of location nodes and $\mathcal{E}$ is the set of edges linking persons to locations, the SEIR model needs to predict, what the numbers of individuals for these populations are in different locations at time step $t$, according to the preceding populations progress, the current mobility network and a set of parameters controlling the probability of transferring from one state of population to the next state.

Formally, we use the superscript $^l$ to represent a specific location $l$ and use the subscript $_t$ to represent the time point $t$, e.g., $S^l_t$ is the number of individuals for state $S$ in location $l$ at time step $t$. 
The number for all populations in location $l$ at time step $t$ is denoted by $N^l_t=S^l_t+E^l_t+I^l_t+R^l_t$, and $S_t=\sum_l S^l_t$ is the population for all locations at time $t$ (the same for $E_t, I_t, R_t$).
We further assume a constant overall population $N=N_t=S_t+E_t+I_t+R_t,\forall t$. There are three sets of parameters $\beta, \kappa$ and $\gamma$ controlling how likely that a person currently at one particular state would transfer to the next state. 
$\beta$ controls the probability of transferring from $S$ to $E$, and $\kappa$ and $\gamma$ are respectively responsible for tranferring from $E$ to $I$ and $I$ to $R$.
$\beta$ can be a set of values, representing the transmission rate for different location categories or cities.
$\kappa$ and $\gamma$ can also be a set of values, representing the 
infection and 
recovery rates for different population groups. 
With a fixed set of parameters and the mobility network, the SEIR model can simulate the disease spread process and predict the numbers of these four populations at each time step for a location. 
The probability of an individual transferring from one state to the next at location $l$ for time step $t$ can be formulated as:
\begin{equation}
  \begin{cases}
  p^l_t(S\to E)=\beta \frac{I^l_t}{N^l_t}\\
  p^l_t(E\to I)=\kappa\\
  p^l_t(I\to R)=\gamma\\
  \end{cases}
  \label{eq1}
\end{equation}
During simulations, the state for each individual is sampled. 
An illustration is shown in Figure \ref{fig:seir}.
The sum of the $L_2$ distances between the observed infections and the predicted infections over all time steps is the simulation error for a particular set of parameters, and the best parameter combination is selected to minimize the simulation error:
\begin{equation}
\Theta^*_\text{SEIR}=\beta^*,\kappa^*,\gamma^*=\arg\min_{\beta,\kappa,\gamma}\sum_t\|I_t-\hat{I}_t\|^2
\label{eq2}
\end{equation}
We use the hat notation $\hat{~}$ to represent the prediction made by the model.
Towards finding the (near) optimal parameter combination in Eq.(\ref{eq2}), the SEIR models needs to run Eq.(\ref{eq1}) over all individuals for $T$ time steps, which gives a total time complexity of $\mathcal{O}(NT)$. 

\section{Modeling the SEIR Process Using Recurrent Nets}
\subsection{Overview}
The core idea of the proposed framework is that, instead of using the brute force grid search strategy to obtain the parameter set that leads to the least $L_2$ distance between predictions and observations, we train a recurrent net to fit the time-series data of SEIR based on a small number of simulations using different sets of parameters.
The trained recurrent net can then be used to directly find the optimal value of  $\Theta^*_\text{SEIR}$, which denotes the parameter set for SEIR that leads to the optimal predictions. 
We employ the widely used Long Short-Term Memory network (LSTM) \cite{hochreiter1997long} as the recurrent net model backbone.

\subsection{Simulation Dataset Construction}
The first step of the proposed framework is to generate smiulation data, which will be used to train the recurrent LSTM net.
We first use different sets of $\Theta_\text{SEIR}$ to perform simulations on the predefined network, as in Eq.(\ref{eq1}). 
We perform $M'$ simulations with $M'$ different sets of 
$\Theta_\text{SEIR}$, denoted by $\{\Theta_\text{SEIR}^{(m')}\}_{m'=1}^{M'}$. 
At each time step $t$, we iterate over all individuals, assign the previous state (i.e., $S$, $E$, $I$ or $R$) of an individual, and sample its current state based on the SEIR model. 
Then, we sum up all individuals belonging to the same states, and obtain $\{S_t^{(m')}, E_t^{(m')}, I_t^{(m')}, R_t^{(m')}\}$ for time step $t$.

\subsection{Training the Recurrent Nets}
Next, we train an recurrent net to fit  $\{S_t^{(m')}, E_t^{(m')}, I_t^{(m')}, R_t^{(m')}\}$.  
Our goal is to use LSTM to predict the values of $\{S_t^{(m')}, E_t^{(m')}, I_t^{(m')}, R_t^{(m')}\}$ for each time step $t$ for each parameter combination in $\{\Theta_\text{SEIR}^{(m')}\}_{m'=1}^{M'}$.
Specifically, 
for each time step $t$, the input to the LSTM model is denoted by $\bm{x}_t^{(m')} = \{S_{t-1}^{(m')}, E_{t-1}^{(m')}, I_{t-1}^{(m')}, R_{t-1}^{(m')}\}$, the populations of the four states at the previous time step.
The model should also be aware of 
the value of 
 $\Theta_\text{SEIR}^{(m')}$ and the mobility network structure at time step $t$ because they directly decide 
 the number of individuals for the four states
  for the next time step.
 To serve this propose, we also feed $\Theta_\text{SEIR}^{(m')}$ as input to the LSTM model at time step $t$.
 For the time-varying mobility network $\mathcal{G}_t$, we map it to a vector representation $\bm{h}_{\mathcal{G}_t}$, which is able to capture its mobility structure and be conveniently fed to the LSTM network.

\paragraph{Mapping $\mathcal{G}_t$ to $\bm{h}_{\mathcal{G}_t}$} We use the DIFFPOOL model \cite{ying2018hierarchical} to map the mobility network $\mathcal{G}_t$ at time step $t$ to its high-dimensional representation $\bm{h}_{\mathcal{G}_t}$.
DIFFPOOL is a differentiable pooling model that can generate hierarchical representations of graphs by progressively clustering nodes into a coarser graph. Specifically, at each layer $l$, DIFFPOOL learns an assignment matrix $\bm{S}^{(l)}\in\mathbb{R}^{n_l\times n_{l+1}}$ to assign each node at layer $l$ to a cluster in the next layer $l+1$, where $n_l$ is the number of nodes (clusters) at layer $l$. Given node embeddings $\bm{Z}^{(l)}\in\mathbb{R}^{n_l\times d}$ and the adjacency matrix $\bm{A}^{(l)}\in\mathbb{R}^{n_l\times n_l}$ at layer $l$, DIFFPOOL generates the new node embeddings $\bm{Z}^{(l+1)}\in\mathbb{R}^{n_{l+1}\times d}$, the new adjacency matrix $\bm{A}^{(l+1)}\in\mathbb{R}^{n_{l+1}\times n_{l+1}}$ and the new assignment matrix $\bm{S}^{(l+1)}\in\mathbb{R}^{n_{l+1}\times n_{l+2}}$ by applying the following equations:
\begin{equation}
\begin{aligned}
    \bm{Z}^{(l+1)}=\text{GNN}_{l,\text{embed}}(\bm{A}^{(l+1)},\bm{X}^{(l+1)})&,~\bm{S}^{(l+1)}=\text{softmax}\left(\text{GNN}_{l,\text{pool}}(\bm{A}^{(l+1)},\bm{X}^{(l+1)})\right)\\
    \bm{X}^{(l+1)}={\bm{S}^{(l)}}^\top\bm{Z}^{(l)}&,~\bm{A}^{(l+1)}={\bm{S}^{(l)}}^\top\bm{A}^{(l)}\bm{S}^{(l)}
\end{aligned}
\end{equation}
$\text{GNN}_{l,\text{embed}}$ and $\text{GNN}_{l,\text{pool}}$ are two distinctly parameterized GNN, and are respectively responsible for generating new embeddings and producing a distribution over next-layer clusters. Setting the number of clusters at the last layer $L$ to 1, DIFFPOOL outputs a single high-dimensional graph representation. We train DIFFPOOL to classify the graph $\mathcal{G}_{t}$ at time step $t$ to the category label $t$, and use the extracted graph representation $\bm{h}_{\mathcal{G}_{t}}$ as input to the LSTM.

\paragraph{Training LSTMs $\Theta_\text{LSTM}$}
Given $\bm{h}^{(m')}_{t-1}$, $\bm{x}_{t}^{(m')}$, $\Theta_\text{SEIR}^{(m')}$ and $\bm{h}_{\mathcal{G}_{t}}$, we are able to obtain the hidden vector representation $\bm{h}_t^{(m')}$ for the time step $t$:
\begin{equation}
\left[
\begin{array}{lr}
\bm{i}_t\\
\bm{f}_t\\
\bm{o}_t\\
\bm{l}_t\\
\end{array}
\right] =
\left[
\begin{array}{c}
\sigma\\
\sigma\\
\sigma\\
\text{tanh}\\
\end{array}
\right]
\bm{W}\cdot
\left[
\begin{array}{c}
\bm{h}_{t-1}^{(m')}\\
\bm{x}_{t}^{(m')}\\
\bm{h}_{\mathcal{G}_{t}} \\
\Theta_\text{SEIR}^{(m')} \\
\end{array}
\right]
\end{equation}
\begin{equation}
\bm{c}_t=\bm{f}_t\cdot \bm{c}_{t-1}+\bm{i}_t\cdot \bm{l}_t
\end{equation}
\begin{equation}
\bm{h}_{t}^{(m')}=\bm{o}_t\cdot \text{tanh}(\bm{c}_t)
\end{equation}
where $\bm{W}_i$, $\bm{W}_f$, $\bm{W}_o$, $\bm{W}_l \in \mathbb{R}^{K\times (2K+4+|\Theta_\text{SEIR}|)}$ where $K$ is the dimensionality of $\bm{h}_{t}^{(m')}$ and $\bm{h}_{\mathcal{G}_{t}}$.
$\bm{h}_{t}^{(m')}$
is then passed to a fully connected layer to obtain $ \bm{\hat{h}}_{t}^{(m')}$, which is mapped to scalars to predict $ \{S_{t+1}^{(m')}, E_{t+1}^{(m')}, I_{t+1}^{(m')}, R_{t+1}^{(m')}\}$: 
\begin{equation}
\begin{aligned}
 \hat{S}_{t+1}^{(m')}  &=  \bm{h}_S^\top \times \bm{\hat{h}}_{t}^{(m')}  \\
  \hat{E}_{t+1}^{(m')} &=  \bm{h}_E^\top \times \bm{\hat{h}}_{t}^{(m')} \\ 
 \hat{I}_{t+1}^{(m')} &=  \bm{h}_I^\top \times \bm{\hat{h}}_{t}^{(m')}  \\
 \hat{R}_{t+1}^{(m')} &=  \bm{h}_R^\top \times \bm{\hat{h}}_{t}^{(m')}  \\
 \end{aligned}
\end{equation}
where $\bm{h}_S$, $\bm{h}_E$, $\bm{h}_I$, $\bm{h}_R\in \mathbb{R}^{K\times 1}$. 
The training objective is  minimizing  the distance between simulation outputs
 $ \{S_{t+1}^{(m')}, E_{t+1}^{(m')}, I_{t+1}^{(m')}, R_{t+1}^{(m')}\}$ and 
 LSTM predictions
 $ \{\hat{S}_{t+1}^{(m')}, \hat{E}_{t+1}^{(m')}, \hat{I}_{t+1}^{(m')}, \hat{R}_{t+1}^{(m')}\}$:
\begin{equation}
\small
\Theta^*_\text{LSTM} = \arg\min_{\Theta_\text{LSTM}} \sum_{t,m'}[ \|S_t^{(m')} - \hat{S}_t^{(m')}\|^2+ \|E_t^{(m')} - \hat{E}_t^{(m')}\|^2+ \|I_t^{(m')} - \hat{I}_t^{(m')}\|^2+ \|R_t^{(m')} - \hat{R}_t^{(m')}\|^2 ] 
\label{lstm}
\end{equation}
Eq. \ref{lstm} can be trained in an end-to-end fashion to obtain optimal  $\Theta_\text{LSTM}$. 
\subsection{Finding Optimal $\Theta_\text{SEIR}$}
The LSTM model with $\Theta^*_\text{LSTM}$ is able to generalize the behavior of the SEIR model with a specific value of  $\Theta_\text{SEIR}$. 
 When training the LSTM to learn $\Theta^*_\text{LSTM}$, $\Theta_\text{SEIR}$ is set to a fixed value of $\Theta_\text{SEIR}^{(m')}$ and fed as input to LSTMs at each time step. 
Due to the fact that $\Theta_\text{SEIR}$ can also be viewed as parameters in LSTMs, i.e., the input to each time step, 
we can fix  $\Theta^*_\text{LSTM}$ and relax  $\Theta_\text{SEIR}$, treating  $\Theta_\text{SEIR}$ as learnable parameters, to minimize Eq.(\ref{eq2}) which we write down here for reference:
\begin{equation}
\Theta^*_\text{SEIR}=\beta^*,\kappa^*,\gamma^*=\arg\min_{\beta,\kappa,\gamma}\frac{1}{T}\sum_t\|I_t-\hat{I}_t\|^2 \tag{2}
\end{equation}
where $\hat{I}_t$ is the output from LSTM, and $\hat{I}_t$ is the observation data rather than the simulation data used to train the LSTM. Eq.(\ref{eq2}) is differentiable with respect to $\Theta_\text{SEIR}$ and can be trained in an end-to-end fashion based on SGD \cite{kiefer1952stochastic,rumelhart1986learning}.
To this end,
we learn the optimal values of $\Theta_\text{SEIR}$ that minimize the $L_2$ distance between predicted infections and observations. 

In the case where we have prior knowledge about the values  $\Theta_\text{SEIR}$, e.g., all values in $\Theta_\text{SEIR}$ should be larger than 0,  
the value of $\beta$ is usually smaller than 0.1 based
 on clinical observations for COVID-19 \cite{chang2021mobility}, 
we can incorporate regularizers as side objectives: 
\begin{equation}
\Theta^*_\text{SEIR}= \arg\min_{\Theta_\text{SEIR}}= \sum_t \frac{1}{T}\|I_t-\hat{I}_t\|^2  + \lambda ||\Theta_\text{SEIR} - \text{prior}(\Theta_\text{SEIR}) ||^2
\label{seir}
\end{equation}
where $\text{prior}(\Theta_\text{SEIR})$ denotes the human prior knowledge regarding the values of $\Theta_\text{SEIR}$, and $\lambda$ controls the trade-off. We will explore the effects of $\text{prior}(\Theta_\text{SEIR})$ and $\lambda$ in experiments.

\section{Experiments}
\subsection{Datasets and Corresponding SEIR Models}
We use two public datasets for evaluations:  the infection network of Covid-19 in China  (Covid-China)  \cite{sun2021analysis,liu2021mobility}, and the infection network of Covid-19 in  the US   (Covid-US) \cite{chang2021mobility}.

{\bf Covid-China} consists of infection networks for 31 provinces in China from Apr 2020 to Feb 2021,
extracted from action tracking reports of Covid-19 patients.
The network of Covid-China consists of 
  two types of nodes: patient and location. 
Time-varying edges
are  
 constructed between a patient node and a location node if the patient visited the location at  time $t$.
Each location takes an attribute from 11 categories of locations: \texttt{households}, \texttt{workplaces}, \texttt{hotels}, \texttt{supermarkets}, \texttt{banks}, \texttt{restaurants}, \texttt{parks},\texttt{barber shops/hairdressers}, 
\texttt{trains}, \texttt{buses}, and \texttt{airplanes}, 
and the economic city tier (first, second or third) that it belongs to. 
A patient node is characterized by features of age (taking the value of children, youths, adults or seniors) and gender (taking the value of male and female). 
Each attribute for gender, age, city-tier and location type is associated with a specific transmission rate $\beta$.  
The transmission rate for a certain person node of gender $s$, age $a$ in location of type $c$ in a city of tier $t$ is
the additive  combination of corresponding $\beta$:
\begin{equation}
\begin{aligned}
\beta(s, a, t, c) = \beta_{\text{s}} + \beta_{\text{a}} + \beta_{\text{t}} + \beta_{\text{c}} \\
\end{aligned}
\end{equation}
$\beta= \{\beta_{\text{s}}, \beta_{\text{a}}, \beta_{\text{t}}, \beta_{\text{c}}\}$, along with $\gamma$ and $\kappa$
are  parameters to learn. 
The network for each city is sliced 
  into consecutive time snippets, with the size of stride set to two weeks.
  For each city, we have daily gold number of infections, public by Chinese CDC.
  These  gold numbers of infections are used to learn $\Theta_{SEIR}$.  
  Snippets without any infection are removed. 
Each time step of each city is labeled with gold number of infections, which is used to train the SEIR parameters.
Snippets are divided to 80\%/10\%/10\% for training, dev and test.

{\bf Covid-US} consists of networks that capture hourly visits from each population group to each location in 10 metro areas in the US. 
The network is extracted from mobility data provided by the SafeGraph application.  
The network consists of two types of nodes: population group and location. 
A time-varying edge with weight $w_{i,j}$ is constructed, if at time $t$, the number of people from population group $i$ visiting location $j$ is $w_{i,j}$.
$w_{i,j}$  is column-normalized. Each location is associated with a location category (e.g., \texttt{full-service restaurants}, \texttt{grocery stores}, etc), 
and each population group is associated with its race and median income.  
For the SEIR model, 
transmissions can happen within groups or  across population groups when two people from two groups visit the same location. 
Simulations are performed at the population group level. 
Each category location $c$ is associated with a specific transmission rate $\beta_c$, which captures the inter-group transmissions across groups in the locations. For each  population group with race $r$ and income decile ($i$), 
the intra-group transmission is set to $\beta = \beta_r + \beta_i$. 
Each area is associated with gold number of infections published by the The New York Times\footnote{\url{https://github. com/nytimes/covid-19-data}}. 
Each snippet consists of the network for a single city and lasts a week. 
We divided snippets  to 80\%/10\%/10\% for training, dev and test. 

\subsection{Experimental Details}
\paragraph{Generating Simulated Data}
We first need to sample $\Theta_\text{SEIR}$. 
We limit the value of each $\beta$ to the scope of [0, 0.1], and we randomly sample its value within the scope. 
For $\kappa$ and $\gamma$, based on previous clinical observations  \cite{kucharski2020early} where 
$\kappa^{-1}$ is around 96 hours and $\gamma^{-1}$ is around 84 hours, we sample $\kappa$ and $\gamma$
from a normal distribution with expectation set to 96 and 84. 
Given a sampled set of $\Theta_\text{SEIR}$, we run simulations on the training datasets to obtain the simulation data, i.e., the number of individuals for all four 
states for each time step.
For each episode, we take $K$ samples of $\Theta_\text{SEIR}$, 
 leading to a total number of $K*|\text{train}|$ training sequences, where $|\text{train}|$ denotes the number of training episodes. 
\paragraph{Learning LSTMs $\Theta_\text{LSTM}$ to Fit Simulated Data}
 We split the simulated data to 90/10 for training and validation. 
We train a three-layer LSTM with residual connections \cite{kaiming2016resnet,kim2017residual} to fit the simulated time-series data using for training, based on Eq.(\ref{lstm}). 
Then size of  hidden states is set to 128. 
The value of batch size is set to 256, and SGD is used for optimization. 
LSTM parameters and embeddings are initialized from a uniform distribution in [-0.08,008]. 
Gradient clipping is adopted by scaling gradients when the norm exceeds a threshold of 1.
Dropout rate, learning rate and the number of training epochs are treated as hyper-parameters to be tuned on the dev set. 

\paragraph{Learning $\Theta_\text{SEIR}$}
We optimize $\Theta_\text{SEIR}$ based on Eq.(\ref{seir}) on the number of gold daily  infections, using the daily reported infections as labels. 
We use AdaGrad \cite{duchi2011adaptive} for optimization. 
Dropout rate, learning rate, the number of training epochs, and the hyper-parameter $\lambda$ 
are tuned on the dev set. 

\begin{table}[t]
    \centering
    \small
    \begin{tabular}{lcccc}\toprule
                            & \multicolumn{2}{c}{\it Covid-China} & \multicolumn{2}{c}{\it Covid-US}\\
                            \cmidrule(r){2-3} \cmidrule(r){4-5}
        {\bf \# Simulations}     & {\bf Vanilla} & {\bf LSTM}          & {\bf Vanilla} & {\bf LSTM}\\\midrule 
        {\bf 20}	&34.2&	22.2&	1870&	1250\\
        {\bf 100}	&30.1&	13.1&	1530&	1130\\
        {\bf 500}  &	18.7&	9.9&	1320&	930\\
        {\bf 1000}&	15.2&	9.2&	1120&	824\\
        {\bf 5000}&	13.5&	8.6&	970&	674\\
    \bottomrule
    \end{tabular}
    \caption{Average square $L_2$ distances for vanilla grid search simulation and recurrent LSTMs on Covid-China and Covid-US}
    \label{tab:baseline}
\end{table}

\subsection{Results}
For baselines, we search the optimal $\Theta_\text{SEIR}$ using vanilla grid search SEIR simulations. 
For each set of $\Theta_\text{SEIR}$, 
simulations are performed on all training episodes, and the parameter set that leads to the minimum $L_2$ loss is selected as the final value.
Suppose that we conduct $K$ explorations for $\Theta_\text{SEIR}$. This means we need to perform $K$ simulations on each training episode. 
The $K$ is here thus comparable to and the same as the $K$ for simulation data generation.  

We report the average of the square of $L_2$ distance between the predicted number  and 
reported number  of infections 
on the test episodes with varying number of simulations $K$ on the test set.
Lower values indicate superior models. 
Results are shown in Table \ref{tab:baseline}.
Observations are as follows: (1) as the number of simulations $K$ increases, 
the performances for both the vanilla model and the proposed  model improve.
This is in accord with our expectations:
for the vanilla model, a larger number of  simulations means that the model is able to explore the search space more thoroughly to obtain the optimal value;
for the proposed LSTM model, the model learns better with more training data and avoids overfitting; 
(2) with the same number of $K$, the proposed LSTM model performs significantly better than the vanilla brute-force search model. This is due to the generalization 
ability of proposed framework: the vanilla model can only select the optimal  parameters from the set it tries, while
the
proposed framework can generalize to the un-tried parameter set;
and (3) notably, the proposed framework is able to achieve comparable performance to the vanilla search model with significantly smaller number of simulations.
Specifically, for Covid-China, the performance obtained with 100 simulations (13.1) is comparable to the vanilla model with 5,000 simulations (13.5);
for Covid-US, the performance obtained with 500 simulations (930) is comparable to the vanilla model with 5,000 simulations (970). 
This further illustrates the superiority of the proposed framework.

\begin{figure}
    \centering
    \includegraphics[scale=0.18]{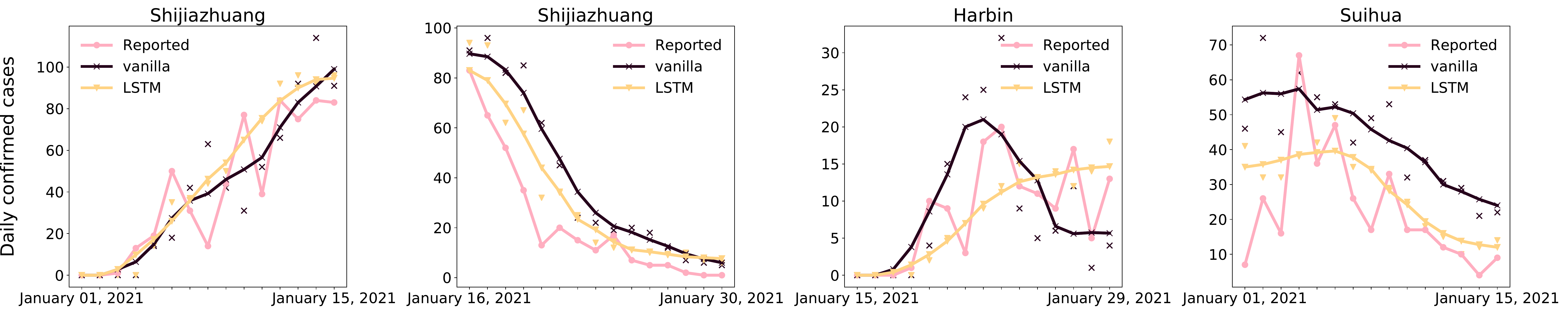}    
    \caption{Daily reported cases, predictions made by the vanilla SEIR model, and the predictions made by LSTM for three cities in China -- Shijiazhuang, Harbin and Suihua. Curves are smoothed by 5-day average. }
    \label{examples}
\end{figure}

Figure \ref{examples} shows simulations performed on the four test episodes in Covid-China dataset using parameters learned from  the proposed model and the vanilla SEIR model .
As can be seen,  the proposed framework offers more accurate predictions. 

\subsection{Ablation Studies}
In this subsection, we explore the effect of different modules, along with hyper-paremeters 
in the proposed framework to explore their influence.

\begin{table}[t]
    \centering
    \begin{tabular}{lcccccccc}\toprule
    $\log_{10}(\lambda)$ & 0 & 1 & 2 & 3 & 4 & 5 & 6 & 7\\\midrule
    {\bf Covid-China} & 10.4 & 9.9 & 9.6 & {\bf 9.4} & 10.1 &11.4 & 13 & 23.5\\ 
    {\bf Covid-US} & 890 & 866 & 861 & 842 & 832 & {\bf 824} & 899 & 1423\\\bottomrule
    \end{tabular}
    \caption{The effect of different values of $\lambda$ on Covid-China and Covid-US. We show the logarithm with base 10 for $\lambda$. The average square $L_2$ distances are reported.}
    \label{tab:lambda}
\end{table}

\paragraph{The effect of $\lambda$}
The hyper-parameter controls the trade-off between observations and external prior knowledge. 
The effect of $\lambda$ is shown in Table \ref{tab:lambda}.
As can be seen, model performance first improves and then declines as the value of $\lambda$ grows.
 Finding the sweep spot for the balance between observations and prior knowledge leads to the best performance. 
 
\begin{table}[t]
  \small
  \centering
  \begin{minipage}[t]{0.3\linewidth}
  \vspace{0pt}
        \centering
        \begin{tabular}{lcc}
        \toprule
        & {\bf CN} & {\bf US}\\\midrule
        {\bf LSTM} & 9.4 & 824\\
        {\bf GRU} & 9.1 & 831\\
        {\bf RNN} & 12.1 & 890\\
        {\bf SRU} & 11.5 & 872\\
        \bottomrule
        \end{tabular}
        \caption{The effect of recurrent structures.}
        \label{tab:structure}
  \end{minipage}%
  \hfill
  \begin{minipage}[t]{0.3\linewidth}
  \vspace{0pt}
      \centering
      \begin{tabular}{lcc}
      \toprule
        & {\bf CN} & {\bf US}\\\midrule
        {\bf Each} & 9.4 & 824\\
        {\bf First} & 13.5 & 1007\\
        {\bf Hadamard} & 9.1 & 821\\
        \bottomrule
      \end{tabular}
      \caption{The effect of using different strategies to incorporate SEIR parameters.}
      \label{tab:incorporate}
  \end{minipage}%
  \hfill
  \begin{minipage}[t]{0.3\linewidth}
  \vspace{0pt}
      \centering
      \begin{tabular}{lcc}
      \toprule
        & {\bf CN} & {\bf US}\\\midrule
        {\bf w/o Graph} & 24.2 & 2531 \\
        {\bf Constant} & 18 & 2013\\
        {\bf Varying} & 9.4 & 824\\
        \bottomrule
      \end{tabular}
      \caption{The effect of graph representations.}
      \label{tab:graph}
  \end{minipage}%
\end{table}

\paragraph{The effect of Recurrent Structures}
We also conduct experiments using other recurrent net structures, including
vanilla recurrent net (RNN), Gated Recurrent Unit (GRU) \cite{cho2014learning}, and Simple Recurrent Units (SRU) \cite{lei2017simple}.
Results for different recurrent structures are shown in Table \ref{tab:structure}.
As can be seen, the LSTM structure performs comparable to GRU (slightly worse than GRU on Covid-China  and slightly better on Covid-US), better than the vanilla recurrent net and SRU.

\paragraph{How to incorporate $\Theta_\text{SEIR}$ into LSTMs}
We explore the effects of different ways to incorporate  $\Theta_\text{SEIR}$ in the LSTM model, including 
the current strategy of (1)
$\Theta_\text{SEIR}$ being concatenated with the input $\bm{x}_t$ at each time step ({\bf Each});
(2) $\Theta_\text{SEIR}$ being incorporated only at the first time step ({\bf First}); and
(3) $\Theta_\text{SEIR}$ element-wise multiplies (Hadamard product)  the input for each time step ({\bf Hadamard}).
For (3), since the dimensionalities of  $\Theta_\text{SEIR}$  and $\bm{x}_t$ are different,  $\bm{x}_t$ is first passed to an FFN, the output of which has the same dimensionality with $\Theta_\text{SEIR}$.
Results for the three strategies are shown in Table \ref{tab:incorporate}.
As can be seen, $\Theta_\text{SEIR}$ incorporated only at the first time step significantly underperforms the strategy that incorporates $\Theta_\text{SEIR}$ at every time step. This is because of 
the gradient vanishing effect of recurrent nets: reminding the model of $\Theta_\text{SEIR}$ at each time leads to better performances. 
The concatenation  strategy obtains comparable performances to the Hadamard product strategy. 

\paragraph{The Effect of Graph Representation}
The time-varying location-population network is captured by the graph representations  through DIFFPOOL. 
This is critical since the number of infections highly relies on the spreading network for each time. 
We explore its necessity by comparing it with other variants: 
(1) no network ({\bf w/o Graph}): where no time-varying graph embedding is incorporated; 
(2) constant graph representation ({\bf Constant}): the graph embedding is not time-varying, where we use the graph embedding of the first time step for all time steps;
(3) time-varying graph representation ({\bf Varying}): the strategy adopted in this work where time-varying representations are incorporated. 
Results are shown in Table \ref{tab:graph}.
As can be seen, when no network information is incorporated, the model nearly fails to learn anything.
Constant networks perform slightly better than no network, but still significantly worse than the time-vary graph representations.
This is in accord with our expectations since the time-varying network decides the number of infections and the disease spread patterns at each time. 

\section{Related Works}
The classical compartmental models simplify the mathematical modeling of disease spread by simulating the population transitions between different states in the disease spread process. 
Since the outbreak of Covid-19, the SEIR model has been widely used to model the Covid-19 spread around the world \cite{estrada2020COVID,godio2020seir,grimm2021extensions} and provide important insights regarding isolation policies \cite{endo2020estimating,carcione2020simulation,chang2021mobility} and vaccine delivery strategies \cite{bubar2021model,ghostine2021extended}.
To achieve a faster simulation speed and better simulation results, recent works have proposed to leverage deep neural networks in place of SEIR models to predict pandemic dynamics over time. For example, \cite{yang2020modified} used the LSTM model to predict the numbers of new infections given the contact statistics and the pre-selected transmission/incubation/recovery/death rates.
\cite{gao2021stan} incorporated sequential network structures and graph attention to predict the number of infections upon a temporal and spatial mobility graph.
These works aim at taking advantage of SEIR models to inform effective strategies in response to the disease spread.

With regard to the estimation of the parameters in SEIR models, a simple approach is to enumerate parameter combinations, run the SEIR model with each combination and select the one with the smallest error. 
An alternative to grid search is to use approximate Bayesian computation (ABC) \cite{sunnaaker2013approximate,saulnier2017inferring,raynal2019abc}, a technique that maintains a small fraction of simulations that are close to the targt statistics in the light of the computed distance. These simulations are treated as posterior distributions of the SEIR parameters, which are then used to infer the optimal parameters.
Another approach to estimating parameters is to view the process of population transitions between states as a problem of ordinary differentiable equations (ODEs) \cite{kermack1927contribution,hethcote2000mathematics,harko2014exact}.
However, ODEs can only give approximate numerical solutions, which could be inaccurate for real-world modeling.

The most relevant work is from \cite{tessmer2018can}, who proposed to make direct use of existing neural networks to predict the basic reproduction number ($R0$),  the number of secondary cases generated by an infectious individual in a fully susceptible host population. This work is different from \cite{tessmer2018can} in that (1) they sought to estimate the basic reproduction number $R0$ and we estimate the parameters in SEIR (or other SEIR variants) models; and more importantly (2) they propose to directly output the number to estimate given inputs, whereas we propose to automatically learn the parameters through neural network gradient descent and back-propagation.
The proposed method can be extended to other fields that require time-consuming simulations to estimate necessary parameters.

\section{Conclusion}
In this work, we propose to transform the original non-differentiable simulation problem of SEIR parameter estimation into a differentiable one by leveraging neural recurrent nets. The recurrent net is first trained to fit a small number of simulation data, and then trained on the observation data to derive the optimal SEIR parameters. This strategy bypasses the needs of time-consuming simulations, and automatically induces the optimal parameters via gradient descent, leading to both accuracy and efficiency gains.

\bibliography{custom}
\bibliographystyle{plainnat}


\end{document}